\documentclass[11pt]{article}
\usepackage[a4paper]{geometry}
\usepackage{clic2016}
\usepackage{times}
\usepackage{url,graphicx,caption,float}
\usepackage{latexsym}
\usepackage{mdframed}
\usepackage{capt-of}
\usepackage[table]{xcolor} 
\usepackage[tracking=true]{microtype}
\usepackage{examples}
\usepackage{wrapfig}

\newcommand\normalstyle{\SetTracking{encoding=*}{0}\lsstyle}

\title{Tracing metaphors in time through self-distance in vector spaces}

\normalstyle

\author{
Marco Del Tredici\\
ILLC, Univ. of Amsterdam\\
Amsterdam, The Netherlands\\
\url{marcodeltredici@gmail.com}\\
\And
Malvina Nissim\\
CLCG, Univ. of Groningen\\
Groningen, The Netherlands\\
\url{m.nissim@rug.nl}\\
\And
Andrea Zaninello\\
Zanichelli editore\\
Bologna, Italy\\
\url{azaninello@zanichelli.it}
}

\date{}

\begin{document}
\maketitle
\begin{abstract}
  \textbf{English.} From a diachronic corpus of Ita\-lian, we build consecutive vector spaces in time and use them to compare a term's cosine similarity to itself in different time spans. We assume that a drop in similarity might be related to the emergence of a metaphorical sense at a given time. 
  Similarity-based observations are matched
  to the actual year when a figurative meaning was documented in a reference dictionary and through manual inspection of  corpus occurrences.
  
\end{abstract}
\begin{abstract-alt}
 \textrm{\bf{Italiano.}} Nel presente esperimento costruiamo spazi vettoriali progressivi nel tempo su un corpus diacronico dell'italiano e calcoliamo la distanza di alcuni termini rispetto a loro stessi in differenti periodi. L'ipotesi \`e che un calo di similitudine possa essere indicativo dell'acquisizione di un significato metaforico. Tale ipotesi \`e valutata attraverso una risorsa lessicografica esterna e l'annotazione manuale dei contesti dei termini nel corpus.
\end{abstract-alt}

\section{Introduction}

It is widely acknowledged that metaphors are pervasive in language use, and that their detection and interpretation are crucial to language processing \cite{group2007mip,turney2011literal,shutova2016design}.

One tricky aspect related to metaphors is their dynamic nature: new metaphors are created all the time. For example, in recent years the Italian term ``talebano'' (`Taliban'), previously only used to refer to the Islamic fundamentalist political movement founded in the Nineties in Afghanistan (Example~\ref{exa:talib-lit}), has come to define more generally someone who is extreme in his or her positions, for example regarding food, use of medicines, and the like (Example~\ref{exa:talib-fig}).\footnote{All of the examples in this paper are from the newspaper \textit{la Repubblica}, see Section~\ref{sec:corpus}.}

\begin{examples}
\item \label{exa:talib-lit} (\textit{lit.}) l'operazione 
[...] ha permesso di arrestare un \textbf{talebano} esperto in esplosivi 
\itemsep 0pt
    \item \label{exa:talib-fig} (\textit{fig.}) [...] 
    senza l'atteso top player, e di un allenatore \textbf{talebano} della tattica
\end{examples}

\noindent If the metaphorical meaning becomes commonly used, it might get recorded in reference dictionaries, too. Indeed, for the case of ``talebano'' the Italian dictionary Zingarelli \cite{zingarelli} has recorded the metaphorical extension (``che (o chi) \`{e} dogmatico, integralista'') in the year 2009, while until then only the literal meaning was included.

Most of the computational work on metaphors has focused on their identification and interpretation using a variety of techniques and models, such as clustering \cite{shutova2013unsupervised}, LDA topic modeling \cite{heintz2013automatic}, tree kernels \cite{hovy2013identifying}, but all from a purely synchronic perspective.\footnote{For a detailed survey on current NLP systems for metaphor modeling see \cite{shutova2016design}.}
The way metaphors develop across time, instead, and whether the shift of a word's literal meaning to a figurative one can be automatically detected and modelled is as of now a little investigated aspect.

As a contribution in this sense, we build on the basic observation that if a metaphorical meaning is acquired by a term at a certain point in time, the context of use of that term will, at least partially, change. 
In this paper we offer a proof of concept of this assumption, based on a selection of terms. 
(Dis)similarity of contexts is measured relying on the distributional semantics approach, and thus on the terms' vector representations, and the existence of a metaphoric shift is derived from the Zingarelli dictionary of Italian.

\section{Approach}
\label{sec:approach}

According to the principle of distributional semantics, the meaning of a word is represented by vectors that encode the contextual information of that word in a corpus \cite{turney2010frequency}. All vectors representing words are included in a distributional semantic space in which similar words are represented by vectors that are close in that space, while different words are distant.

We rely on the intuition that if a term develops a metaphoric sense, its contexts of occurrence will start to differ, at least partially, from those observed for the very same term at the time the metaphorical meaning had not emerged yet. This implies that detecting a distance in space across time could be indicative of a meaning shift.
Hence, instead of comparing different terms synchronically, we focus on their \textit{self-distance} across time, thus tracing their diachronic evolution of meaning.

Practically, we train vector representations of words in consecutive time spans, and compare such representations to one another, for a set of pilot terms. As a default, a term is expected to exhibit a vector representation roughly similar to itself across time. If we observe a drop in similarity between vectors in consecutive spaces, we can hypothesise the emergence of a new sense for this term, potentially metaphoric.

By using the information recorded for the selected terms in a reference dictionary for the Italian language, we observe whether there is some correspondence between the observed similarity drop, if present, and the time of inclusion of a figurative sense. Finally, for each year cluster, we manually inspect the occurrences of our target terms in order to see if changes of use can be observed. 

We are aware of the fact that changes in distance of a word to itself across time might be triggered by phenomena other than the rise of a metaphoric shift. Indeed, especially for polysemous words, extra-linguistic factors could cause the dominance of one sense over the others at a given time. In a larger-scale, bottom-up approach to detect metaphorical shifts, this would need to be properly accounted for. In the context of this proof-of-concept, we control for this factor by choosing words that are not or are minimally polysemous (see Section~\ref{sec:zingarelli}).

\section{Related Work}
\label{sec:related}

The automatic modelling of diachronic shift of meaning has been investigated employing several different techniques. Among these, most recently, Latent Semantic Analysis \cite{sagi2011tracing,jatowt2014framework}, topic clustering \cite{wijaya2011understanding} and dynamic topic modeling \cite{frermann2016bayesian}.
Vector representations for diachronic shift of meaning have been used by \newcite{gulordava2011distributional}, with a simple co-occurence matrix of target words and context terms. \newcite{jatowt2014framework} and  \newcite{xu:2015} experimented both with a bag-of-words approach and a more linguistically motivated representation that also captures 
the relative position of lexical items in relation to the target word. 

Recently, Word Embeddings (\newcite{mikolov2013distributed}, see also Section~\ref{sec:model}) have been used to investigate diachronic meaning shifts: 
vectors are usually created independently for each time span and then mapped from one year to another via a transformation matrix, thus leveraging the stability of the relative positions of vectors in different spaces \cite{kulkarni2015statistically,zhang2015omnia,hamilton2016diachronic}.

An alternative approach, which we also adopt -- with a slight change -- in our work, is introduced by \newcite{kim2014temporal}, who propose a simple but effective methodology to make vectors trained on different corpora directly comparable: embeddings created for year~$y$ are used to initialise the vectors for year $y+1$. The process is  progressively applied to all time spans.

\begin{table*}[ht]
\centering
\caption{\label{tab:terms}Selected terms. \textbf{a-date} = first attested; \textbf{d-date} = decision date for extended meaning to be included in dictionary; \textbf{i-date} = actual inclusion date in Zingarelli for extended meaning.}
\begin{tabular}{llp{7cm}ccc}
\hline
\textbf{term}          & \textbf{literal}    & \textbf{figurative} & \textbf{a-date}  &  \textbf{d-date} & \textbf{i-date} \\ \hline
implosione    &  implosion   &  cedimento, tracollo improvviso
(collapse)   & 1932    &  2013   & 2015 \\
& & & \vspace*{-.4cm} \\
kamikaze      &  kamikaze    & chi compie un'impresa rischiosa o destinata al fallimento (daredevil, reckless)   & 1944    & 2007    & 2009 \\
& & & \vspace*{-.4cm} \\
rottamatore   & dismantler&  nel linguaggio giornalistico e della politica, chi si propone di allontanare e sostituire un gruppo dirigente considerato antiquato (new broom)
   & 1990    & 2012    & 2014 \\
& & & \vspace*{-.4cm} \\
talebano      & Taliban       &   che (o chi) \`{e} dogmatico, integralista (hard-liner, extremist)
  & 1995    & 2007    & 2009 \\
& & & \vspace*{-.4cm} \\
tsunami       & tsunami       &  evento che determina lo sconvolgimento di un assetto costituito
(devastation, havoc)   & 1907    & 2008    & 2010 \\ 
\hline
\end{tabular}
\end{table*}

\section{Experiment}
\label{sec:experiment}

Following the approach described in Section~\ref{sec:approach}, we selected a small set of pilot terms from a lexicographic reference, and observed their space development across time, on a diachronic corpus for Italian that we collected for this purpose. Due to the absence of datasets in which words are annotated for meaning change, a qualitative analysis of a set of hand-selected words like the one we propose has established itself as a common evaluation method in previous work on diachronic meaning change \cite{frermann2016bayesian}.

\subsection{Lexicographic reference and term selection}
\label{sec:zingarelli}

The Zingarelli dictionary is a reference dictionary for the Italian language, updated and published every year, both in digital and paper version. The dictionary is traditionally dated one year ahead of the year it is published, hence the Zingarelli 2017 is published in June 2016, and it refers to decisions about new words and new meanings (including metaphorical ones) made up until December 2015.

We analysed the behaviour of a small set of terms extracted from the dictionary. We searched the 2017~edition to extract nouns that record a figurative meaning, limiting our search to words whose first occurrence is recorded in the 20th or 21st century. Newly born words (including borrowings) are more likely to show a meaning shift in the time span considered in our search (1984-2015) than older words (especially if derived directly form Latin, where the figurative meaning was also originally highly available, so probably arisen earlier).  Out of a total of 447 hits, five target words were chosen for this pilot study. They are reported in Table~\ref{tab:terms} together with relevant information. 

In order to minimise (at least in the context of this experiment) the influence of polysemy in the observable similarity distance across years, we verified that the selected terms are not polysemous, or minimally so. For the words ``rottamatore'', ``talebano'', and ``tsunami'', the Zingarelli records one sense only. 
For the word ``implosione'' three senses in total are recorded, two of which are however technical language, in the fields of linguistics (phonology) and psychology, and we assume will not be used much in newswire. 
For ``kamikaze'' the Zingarelli records one meaning only (Japanese pilot) to which is associated the extended sense of someone who kills himself in a terrorist attack; in our corpus the extended meaning is clearly the primary one, and the figurative sense that we consider is derived from it (see also Section~\ref{sec:discussion}).

\subsection{Corpus}
\label{sec:corpus}

We created a diachronic corpus of approximately 60 millions tokens by collecting articles from the Italian newspaper \textit{la Repubblica} from 1984 (the first year for which data is available digitally) to 2015. All texts were tokenised and lowercased. 
Because we are interested in how a term's context changes over time, we had to determine time-spans for our corpus, and we settled on two-year blocks, for a total of 16 time spans, the first one being 1984-1985 and the last 2014-2015. These subcorpora are used to train consecutive vector space models.

\subsection{Model}
\label{sec:model}

We implemented vector representations using the skip-gram architecture introduced by \newcite{mikolov2013distributed}. 
Such representations (Word Embeddings) are low dimensional, dense and real-valued vectors that have been proved to preserve syntactic and semantic information in several NLP tasks \cite{baroni:14}.

Vectors created on different corpora cannot be directly compared, since every semantic space implements arbitrary orthogonal transformations and hence there is no direct correspondence between word vectors in different semantic spaces \cite{zhang2015omnia}. This would hold true also for our data, since we create a different corpus for each time span.
Therefore, in order to create comparable vector representations for each word in any time span, we adopt the methodology introduced by \newcite{kim2014temporal} (see Section~\ref{sec:related}), slightly modifying it. While \newcite{kim2014temporal} use vectors of span $y$ to initialise the vectors for year $y+1$, we do the opposite, i.e. we start with 2014-15, and use those vectors to initialise the 2012-13 time span, and thus backwards until 1984-85. 

This methodological choice is due to the fact that the majority of the words in the set we considered for this experiment (included the selected target words, see 4.1) have few or no occurrences in the first time spans of the corpus: for example, ``rottamatore" and ``talebano" occur for the first time in 96/97. Indeed, using \newcite{kim2014temporal}'s original approach, which we implemented in a preliminary experiment, the vectors for these words were correctly initialised, but were basically random vectors with no meaningful information. Conversely, our \textit{reverse} setting, while still offering the same opportunity to trace shifts of meaning across time, allows to initialise all target words on a time span (14/15) in which they occur a number of times sufficient to create a more stable, meaningful representation.

Using the \texttt{gensim} library \cite{gensim}, we trained the models with the following parameters: window size of 5, learning rate of 0.01 and dimensionality of 200. We filtered out words with frequency lower than 5 occurrences. The vocabulary was initialised over the whole dataset.

\begin{figure*}[ht]
\includegraphics[width=17cm,height=21\baselineskip]{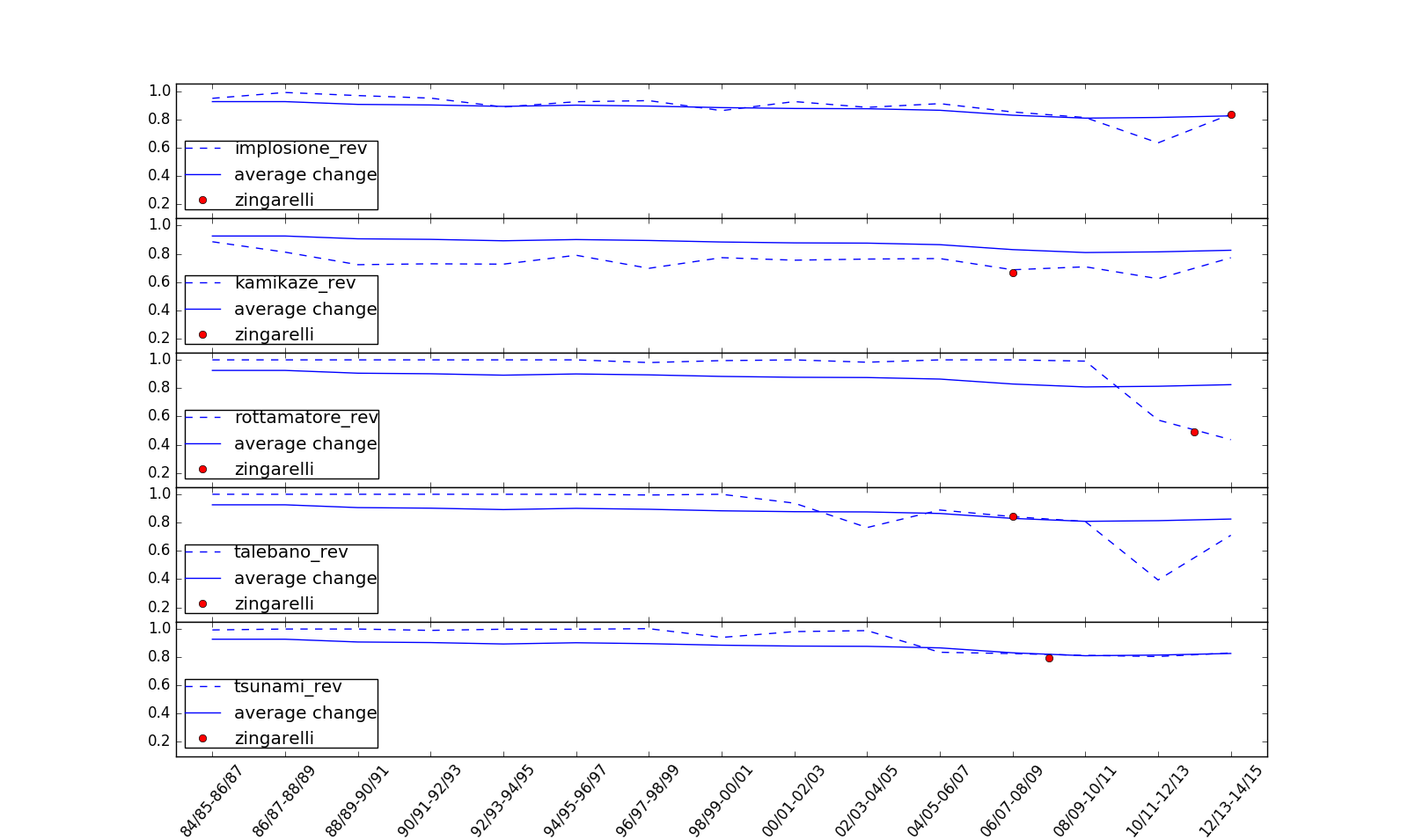}
\centering
\caption{Cosine similarity values across time spans for target words (dotted line), average similarity of nouns (solid line) and date of insertion of metaphorical meaning in the Zingarelli dictionary (red dot).\label{fig:plot}}
\end{figure*}

\subsection{Results and discussion}
\label{sec:discussion}

Figure~\ref{fig:plot} shows the similarity values for one time span to the next (dotted line), together with the average shift of meaning of a subset of 5000 nouns randomly selected (solid line).
While we cannot draw any statistically significant conclusions from such little data, we aim at potentially observing patterns of shift of meaning through change of vector representations that could be used for developing predictive metrics of metaphorical shifts in time.

We interpret the results of our models according to (i) information in the Zingarelli dictionary and (ii) a manual inspection of the context of use of our target words in the corpus.

For (i), we verify if, for a given term, an observable correlation exists between  changes in its vector representations and the insertion of a figurative sense in the dictionary. Results show that such a correlation exists for ``talebano'', ``rottamatore'', and ``tsunami''. For these words a drop in cosine similarity can be observed between three and five years before the insertion of the figurative meaning in the dictionary. This fits well with the timing for new meanings to be recorded in lexicographic resources (see Section~\ref{sec:zingarelli}). 
The nouns ``kamikaze'' and ``implosione'', instead, show a more stable evolution of meaning in time, with no clear drop in cosine similarity, and thus no evident correlation between changes in vector representations and insertion of a figurative meaning in dictionary.

For (ii), we manually inspected the contexts in which target terms occur in the the corpus as literal or metaphoric, in order to check if some relevant change in words usage could be observed in correspondence to drops in cosine similarity between time spans. 

``Tsunami'' occurs 27 times between 84/85 and 02/03: in 88.9\% of the cases the word is used literally, with only 3 metaphorical uses in 98/99 (mirrored in a slight drop in cosine similarity). Of the 930 occurrences from 04/05 to 14/15, only 59.1\% are literal. In Figure~\ref{fig:plot} we can observe a major drop in cosine similarity exactly between 04/05 and 06/06. 

``Rottamatore'' occurs 4 times between 84/85 and 08/09, always used literally. From 10/11 on, there are 156 occurrences, all metaphorical. Thus, the drop corresponds to change in usage here too. 

``Talebano'' occurs 12 times between 84/85 and 02/03, with 83.3\% of literal usage. Once again, the drop in cosine coincides with the time span in which the term started to be used metaphorically: between 02/03 and 08/09 40\% of the occurrences of ``talebano'' are metaphorical. Then, another relevant drop is observed between 08/09 and 10/11, and this is due to the sudden return of the literal usage of this word (86.1\%), which continues also in the following years.

As already noticed, ``kamikaze'' and ``implosione'' do not seem to undergo a clear shift. As for the former, the analysis of its contexts of use reveals that indeed it is not possible to clearly identify, in our corpus, when exactly the term started to be used metaphorically: of the 25 occurrences of ``kamikaze'' in 84/85, 32\% are metaphorical. This trend is fairly constant, and it explains why the vector representation of ``kamikaze'', which from the very beginning conflates literal and metaphorical usages, is stable in time. There is only a relevant change starting from 10/11: from this period onwards, the metaphorical use decreases, and almost all the occurrences are literal.\footnote{Interestingly, this increase of literal usage is observed in the same years also for ``talebano'', a term that is semantically related to ``kamikaze''. This observation would require further investigation in connection with the socio-political events of those time spans.} Accordingly, this almost exclusively return to the literal meaning corresponds to a slight increase in cosine similarity between the two last time spans.

``Implosione'' occurs 433 times overall and in 92.4\% of them is used  metaphorically, but in few and specific contexts. A metaphorical, quite specific, sense of ``implosione'' is thus the main sense for this term in our corpus, and this is why we observe, on average, a high similarity across time spans. There is only a small drop between 10/11 and 12/13, when the word started to be used in the context of the economical crisis (``l'implosione dell'euro''). 

To sum up, both ``kamikaze'' and ``implosione'' show a similar stable behaviour in time, with only small drops. However, while for ``kamikaze'' such stability is due to a relatively constant ratio between literal and metaphorical meanings, in the case of ``implosione'' the observed stability is given by the constant predominance of the metaphorical sense across all the time spans.

\section{Conclusion and future work}

This work was meant as an exploration of the assumption that 
the emergence of the metaphorical use of a term might be mirrored in changes in cosine similarity of the term to itself across time. Such assumption has been partially confirmed by the comparison to the Zingarelli dictionary, while we found a more robust evidence when inspecting the terms' contexts of use manually.

Future work will stem from methodology and observations discussed here. Specifically, we plan to investigate further several  aspects of this initial work, including the relation between changes in cosine similarity and frequency of use of a word: to which extent a change of the former relates to an increase of the latter? Mostly though, we plan to run experiments on larger sets of words with the aim to consolidate and then further exploit the mainly qualitative observations reported here towards the development of reliable predictive metrics which can serve to detect the emergence of shifts automatically, in a completely bottom-up fashion.

\section*{Acknowledgments}

Malvina Nissim would like to thank the ILC-CNR ItaliaNLP~Lab for their hospitality while working on this project. We are also grateful to the anonymous reviewers who provided insightful comments that  doubtlessly contributed to improve this paper.

\end{document}